\documentclass{bmvc2k}

\usepackage{amssymb}
\usepackage{yfonts}
\usepackage{graphicx}

\usepackage[normalem]{ulem} 
\useunder{\uline}{\ul}{}

\usepackage{amsmath}
\usepackage{bbold}
\usepackage{multirow}
\usepackage{booktabs}

\usepackage[normalem]{ulem}
\useunder{\uline}{\ul}{}

\newcommand{\ie}{{\emph{i.e.}}}

\title{Split Matching for Inductive Zero-shot Semantic Segmentation}

\addauthor{Jialei Chen}{jialeichen2021@gmail.com}{1}
\addauthor{Xu Zheng}{zhengxu128@gmail.com}{2,3}
\addauthor{Dongyue Li}{yzly0603@gmail.com}{1}
\addauthor{Chong Yi}{yc.simplerush@gmail.com}{1}
\addauthor{Seigo Ito}{iseigo@vislab.is.i.nagoya-u.ac.jp}{1}
\addauthor{Danda Pani Paudel}{danda.paudel@insait.ai}{3}
\addauthor{Luc Van Gool}{luc.vangool@insait.ai}{3}
\addauthor{Hiroshi Murase}{murase@nagoya-u.jp}{1}
\addauthor{Daisuke Deguchi}{ddeguchi@nagoya-u.jp}{1}

\addinstitution{
 Graduate School of Informatics, Nagoya University (NU), Nagoya, Aichi, Japan
}
\addinstitution{
 Artificial Intelligence Thrust, The Hong Kong University of Science and Technology (Guangzhou) (HKUST (GZ)), Guangzhou, Guangdong, China
}
\addinstitution{
 Institute for Computer Science, Artificial Intelligence and Technology (INSAIT), Sofia University, St. Kliment Ohridski, Bulgaria
}

\runninghead{Split Matching for Inductive Zero-shot Semantic Segmentation}{}

\def\eg{\emph{e.g}\bmvaOneDot}

\begin{document}

\maketitle

\begin{abstract}
Zero-shot Semantic Segmentation (ZSS) targets the segmentation of unseen classes, \ie, classes not annotated during training. While fine-tuned vision-language models show promise, they often overfit to seen classes due to the lack of supervision. Query-based methods offer strong potential by enabling object localization without explicit labels, but conventional approaches assume full supervision and thus tend to misclassify unseen classes as background in ZSS settings. To address this issue, we propose \textbf{Split Matching} (SM), a novel assignment strategy that decouples Hungarian matching into two components: one for seen classes in annotated regions and another for latent classes in unannotated regions (referred to as unseen candidates). Specifically, we split the queries into seen and candidate queries, enabling each to be optimized independently according to its available supervision. To discover unseen candidates, we cluster CLIP dense features to generate pseudo masks and extract region-level embeddings using CLS tokens. Matching is then conducted separately for the two groups based on both class and mask similarity. Additionally, we introduce a \textbf{Multi-scale Feature Enhancement} (MFE) module that refines decoder features through residual multi-scale aggregation, improving the model’s ability to capture spatial details across resolutions. Besides, we also introduce a \textbf{Random Query} (RQ) that injects a few randomly initialized queries during inference. These queries act as unbiased probes, enriching the density and diversity of the query space. Our method is the first to introduce decoupled Hungarian matching under the inductive ZSS setting, and achieves 0.8\% and 1.1\% higher hIoU on two ZSS benchmarks.
\end{abstract}

\section{Introduction}



Semantic segmentation~\cite{FCN, deeplab, frozen,mask2former,maskformer,zheng2025distilling} serves as a fundamental task for computer vision. Existing approaches can be broadly classified into feature-based and query-based methods. Feature-based \cite{FCN, deeplab, frozen} methods treat semantic segmentation as a per-pixel classification problem, where dense features extracted from the backbone are directly finetuned for pixel-wise label prediction. In contrast, query-based segmentors \cite{mask2former,maskformer} employ a set of discrete, learnable vectors, referred to as queries, to jointly predict class labels and class-agnostic masks. These queries are passed through a transformer decoder to produce class scores and interact with dense backbone features to generate the corresponding masks.

However, achieving high performance in semantic segmentation typically requires large-scale datasets with pixel-level annotations \cite{cityscapes}, which are costly to obtain. To reduce annotation demands, Zero-shot Semantic Segmentation (ZSS) \cite{zs3, zegclip, coach, concat} has emerged, aiming to segment unseen classes which are not annotated during training but must be segmented at test time, by transferring knowledge from seen classes, \ie, classes with available training annotations. Recent advances in Vision-Language Models (VLMs) \cite{clip,recognizeanything} drives ZSS by enabling the transfer of vision-language alignment to segmentation tasks \cite{zegformer, zegclip, DeOP}. Existing methods typically adapt CLIP via adapters \cite{adapter} or prompts \cite{visualprompttuning} to handle seen classes, while relying on CLIP’s generalization ability for unseen ones. However, such strategies typically rely on CLIP’s zero-shot capability for unseen classes without additional training signals, which leads to overfitting to seen classes and poor generalization on unseen classes.

\begin{figure}[t]
\begin{center}
\includegraphics[width=0.9\linewidth]{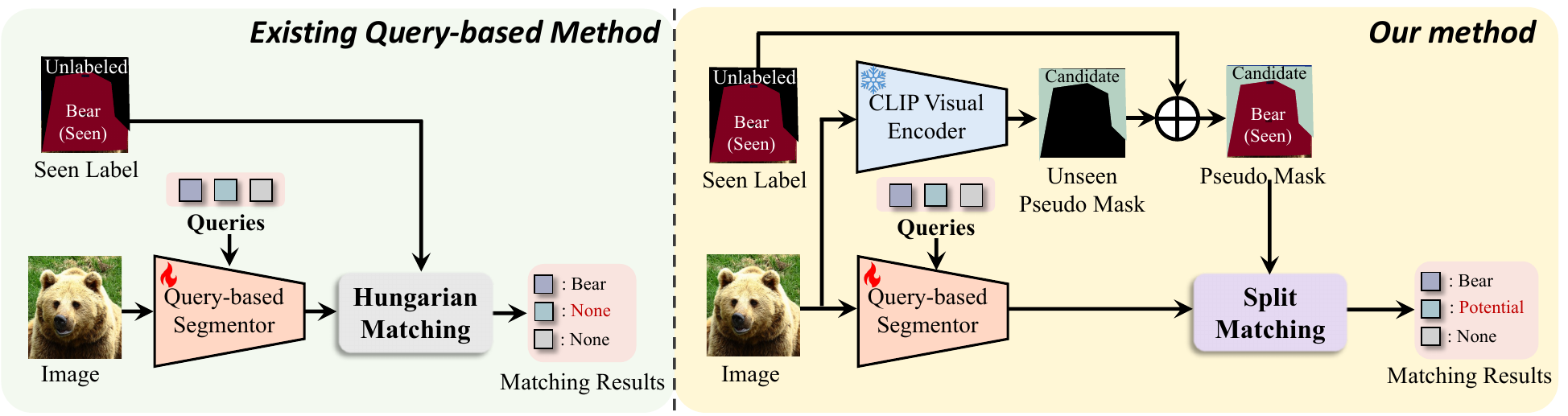}
\end{center}
\vspace{-0.25in}
\caption{Comparisons between existing query-based segmentation models that fail to match the unseen candidates and the proposed split matching.}
\label{fig:teaser}
\vspace{-0.2in}
\end{figure}

To mitigate overfitting to seen classes, we explore query-based segmentation as a more effective solution. Unlike feature-based methods that classify pixels independently, query-based approaches treat each object as a learnable query and perform mask-level classification, enabling better object-level reasoning and localization without explicit supervision. However, despite their ability to localize potential objects, existing query-based methods struggle in ZSS. Without annotations for unseen classes during training, queries for novel objects are often misclassified as seen classes or background. As shown on the left of Fig.~\ref{fig:teaser}, the imbalance biases the model to seen classes and prevents it from learning useful representations for unseen ones, limiting the full potential of query-based segmentation in ZSS.

To improve the ability of query-based methods to segment unseen classes, we propose a method called Split Matching (SM).
Specifically, we divide the queries into two groups: \textbf{\textit{seen queries}}, which segment annotated seen classes, and \textbf{\textit{candidate queries}}, which target latent classes in unannotated regions (referred to as unseen candidates).
We first apply multi-scale K-Means clustering~\cite{coach} on CLIP dense features to generate pseudo masks that localize unseen candidates.
The corresponding image patches are then cropped and fed into the frozen CLIP visual encoder to obtain CLS tokens, which are subsequently fused with semantic embeddings to form joint class embedding.
We compute class similarity between the joint embeddings and query predictions, and measure mask similarity using the pseudo masks.
Hungarian matching is then applied separately to the seen and candidate queries based on the combined similarities. Although SM facilitates the adaptation of query-based approaches to zero-shot segmentation, the key and value features in transformer decoders, responsible for associating queries with relevant image regions, remain suboptimal. These features often lack sufficient spatial granularity and discriminative power, causing noisy query-to-region associations and reduced matching accuracy. Therefore, we introduce a Multi-scale Feature Enhancement (MFE). It enhances the visual feature with multi-scale context and applies spatial normalization, refining the transformer decoder’s key and value through residual multi-scale aggregation. Moreover, we propose a Random Query (RQ) strategy that introduces a small number of randomly initialized queries at inference time. Unlike trained queries, these random queries serve as unbiased exploratory probes that complement the existing query set. By diversifying the query space and increasing its sampling density, RQ reduces blind spots and encourages the decoder to discover additional object regions that would otherwise be missed.

Different from existing methods that rely on dense features extracted from CLIP~\cite{zegclip,coach} and struggle to optimize unannotated regions due to the absence of explicit supervision, our method introduces Split Matching (SM), which separates queries into seen and candidate groups, enabling targeted label assignment even in unannotated areas. This design effectively mitigates the common issue of misclassifying unseen objects as background, a key limitation of prior methods. Moreover, unlike open-vocabulary segmentation approaches~\cite{convolutionsdie,sideadapter} that perform Hungarian matching with fully annotated data, our method operates entirely under the zero-shot setting, without requiring any pixel-level labels for unseen classes. To the best of our knowledge, we are the first to explicitly separate seen and candidate queries via Hungarian matching under the inductive ZSS. To summarize, our contributions are:
	
    •	We propose Split Matching (SM), a novel query-based assignment framework for zero-shot segmentation. SM explicitly separates seen and unseen queries and matches them independently using pseudo masks derived from CLIP dense features.
	
    •	We introduce a Multi-scale Feature Enhancement (MFE) module, which enriches decoder features via residual multi-scale fusion. Additionally, we propose a Random Query (RQ) strategy that increases query density at inference time to uncover more latent objects.
	
    •	Our method achieves 0.8\% and 1.1\% higher hIoU on PASCAL VOC and COCO-Stuff benchmarks under the zero-shot setting.
\vspace{-0.2in}
\section{Related Works}

\noindent \textbf{Semantic Segmentation} assigns a class label to each pixel in an image. Traditional CNN-based methods~\cite{FCN,deeplab,deeplabev3} enhance per-pixel classification through dilated convolutions and multi-scale context aggregation but struggle to capture long-range dependencies. With the rise of Vision Transformers~\cite{vit}, encoder-based models~\cite{segformer,segnext} have shown strong capabilities in modeling global context. Inspired by DETR~\cite{detr}, query-based frameworks such as MaskFormer~\cite{maskformer} and Mask2Former~\cite{mask2former} reformulate segmentation as set prediction using object queries and Hungarian matching. Although these models achieve impressive results under full supervision, they are not directly applicable to zero-shot segmentation (ZSS) due to their reliance on complete annotations and inherent bias toward seen classes. To address this, we propose Split Matching, the first method to explicitly separate seen and candidate queries under the inductive ZSS setting. By decoupling the matching process for annotated and unannotated regions, our approach enables targeted supervision and significantly improves generalization to unseen classes, even in the absence of pixel-level annotations.

\noindent \textbf{Zero-shot and Open-Vocabulary Segmentation.} Although semantic segmentation has made remarkable progress, it still relies heavily on large-scale pixel-level annotations, which are expensive and time-consuming to obtain. To alleviate this, two related directions have emerged: zero-shot segmentation (ZSS) \cite{zegformer,zegclip,DeOP,coach,concat,otseg} and open-vocabulary segmentation (OVS) \cite{sideadapter,convolutionsdie,clipself,maskclip}. Both paradigms aim to improve generalization by leveraging large-scale vision-language models such as CLIP~\cite{clip}, either by introducing lightweight adapters~\cite{adapter,semanticmatter} or designing task-specific visual prompts~\cite{visualprompttuning}. Despite this shared goal, their setups differ fundamentally. ZSS is defined under a partially labeled training regime, where only a subset of classes is annotated and the rest are marked as ignored. Evaluation is conducted on the same domain, including both seen and unseen classes. OVS, on the other hand, assumes access to fully labeled training data and evaluates on datasets with novel classes and distribution shifts. While OVS benefits from full supervision and can leverage existing segmentation architectures, ZSS faces the challenge of incomplete supervision. Unannotated regions corresponding to unseen classes are often treated as background during training, making it difficult for models to learn discriminative representations for unseen concepts. To tackle this issue, we propose \textbf{Split Matching (SM)}, a label assignment strategy tailored for query-based models in the zero-shot setting. SM dynamically aligns predicted queries with pseudo labels derived from external vision-language features, enabling the discovery of unseen objects even in the absence of ground-truth annotations.


\begin{figure*}[t]
\begin{center}
\includegraphics[width=0.8\linewidth]{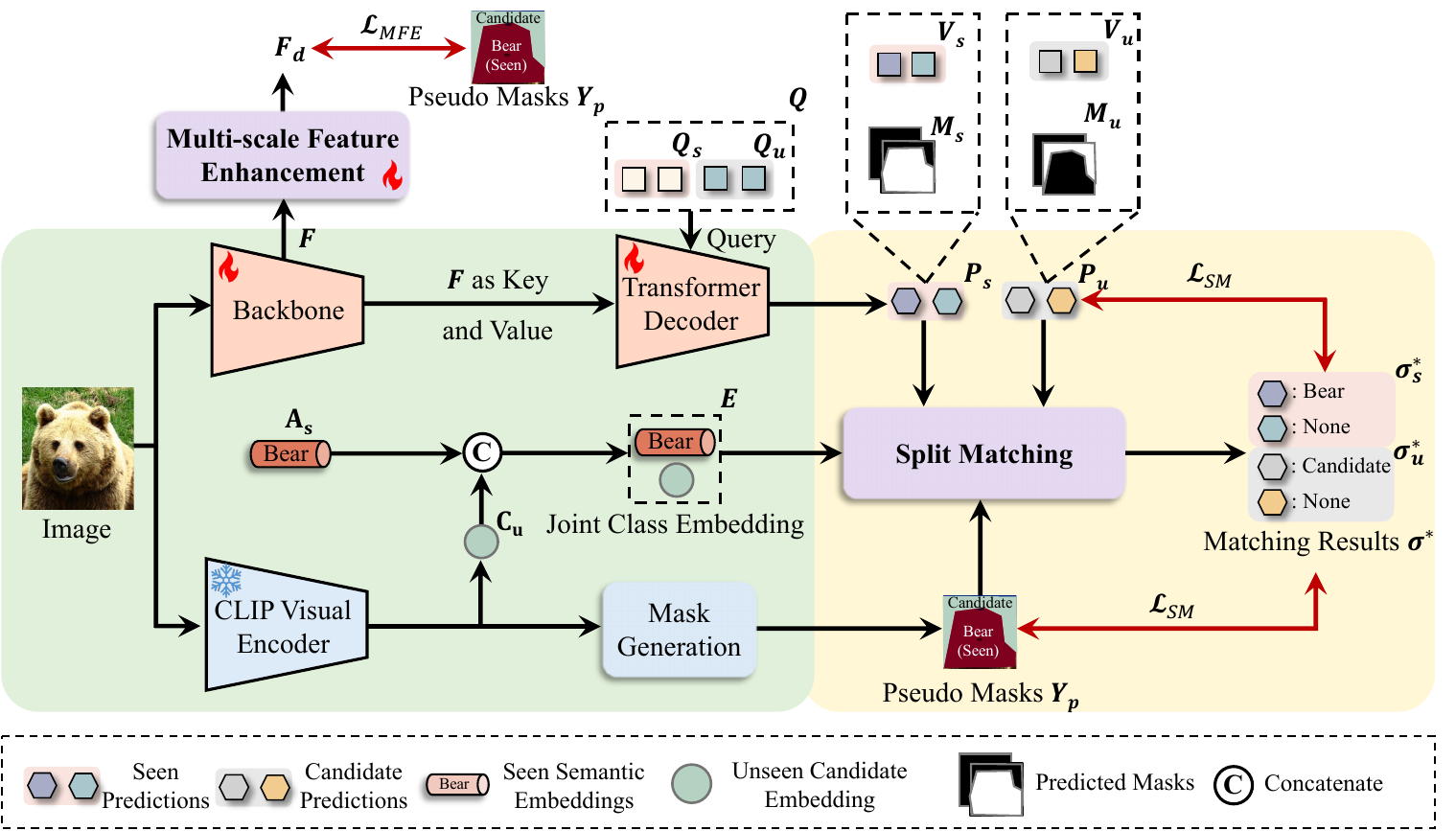}
\end{center}
\vspace{-0.2in}
\caption{Training pipeline overview of the proposed method. During training, input images are fed into a trainable backbone to extract dense features $\textbf{F}$, while a frozen CLIP encoder provides CLIP features. These are used to generate class embeddings $\textbf{C}_u$ and pseudo masks $\textbf{Y}_p$ for both seen and unseen classes. The dense features serve as keys and values in a transformer decoder, which interacts with queries $\textbf{Q}$ to predict masks and query features. These are aligned with pseudo masks via the Split Matching. A Multi-scale Feature Enhancement module further refines $\textbf{F}$ with $\mathcal{L}_{MFE}$.}
\label{fig:main}
\vspace{-0.3in}
\end{figure*}

\section{Methods}
\subsection{Preliminaries}
\noindent \textbf{Task Definition.} ZSS aims to segment both seen and unseen classes without annotations for unseen classes during training. Formally, let $\mathcal{D} = \{\textbf{X}^i, \textbf{Y}_s^i\}_{i=1}^M$ represent a dataset of images $\textbf{X}$ and their pixel-level annotations $\textbf{Y}_s$ for seen classes, and $M$ is the dataset size. Additionally, let $\textbf{A} \in \mathbb{R}^{N \times C}$ denote the semantic (text) embeddings of all classes, divided into seen $\textbf{A}_s \in \mathbb{R}^{N_s \times C}$ and unseen $\textbf{A}_u \in \mathbb{R}^{N_u \times C}$, such that $(\textbf{A}_s \cap \textbf{A}_u = \varnothing)$ and $N_s + N_u = N$ where $C$ indicates the channel number and $N$ is the number of classes in the dataset. Our method applies the \textit{Inductive settings}, where unseen embeddings $\textbf{A}_u$ are inaccessible during training and evaluates model performance on seen and unseen classes during inference. Meanwhile, all training images are preserved during training, and regions corresponding to unseen classes are consistently labeled as ``ignored’’, ensuring that no unseen information is used. Although our method utilizes features from unannotated regions, it strictly adheres to the inductive zero-shot setting.
At the beginning of training, all unannotated regions are uniformly treated as `ignored' without introducing any class-specific supervision or bias.
Unseen candidates are identified in a fully self-supervised manner, without any assumptions regarding the number, identity, or distribution of unseen classes.
Therefore, the model remains agnostic to unseen classes during training, ensuring that no unseen-related information is leaked.

\noindent \textbf{Method Overview}. Our core idea is to mitigate seen-class bias from incomplete annotations by decoupling the optimization of seen and unannotated regions, allowing better discovery of unseen classes without harming seen-class performance. As shown in Fig.\ref{fig:main}, we divide the query space into \textbf{seen queries} $\mathbf{Q}_s$ and \textbf{candidate queries} $\mathbf{Q}_u$, responsible for segmenting annotated seen classes and unseen candidates, respectively. Given an input image, a trainable backbone extracts dense features, which interact with a set of randomly initialized queries through a transformer decoder. These queries are split into $\mathbf{Q}_s$ and $\mathbf{Q}_u$, with the latter guided by pseudo masks and class embeddings for unseen candidates $\mathbf{C}_u$ derived from a frozen CLIP encoder. We then propose \textbf{Split Matching} (Sec.\ref{sec:SM}) to assign labels to both query types via similarity with model outputs, pseudo masks, and class embeddings. Additionally, a \textbf{Multi-scale Feature Enhancement} module (Sec.~\ref{sec:MFE}) is introduced to refine backbone features, and a \textbf{Random Query (RQ)} strategy (Sec. \ref{sec:train}) is introduced during inference.

\subsection{Split Matching (SM)} \label{sec:SM}
Hungarian matching requires splitting ground-truth into class labels and corresponding class-agnostic masks, followed by assigning each ground-truth instance to a query based on similarity. However, in ZSS, unannotated regions lack ground-truth labels, making it impossible to directly apply Hungarian matching, which relies on full supervision. To overcome this, Split Matching (SM) generates pseudo masks for latent classes in unannotated areas and assigns them to candidate queries, while optimizing seen and candidate queries separately.




\begin{figure}[t]
\centering
\begin{minipage}[b]{0.47\linewidth}
  \centering
  \includegraphics[width=\linewidth]{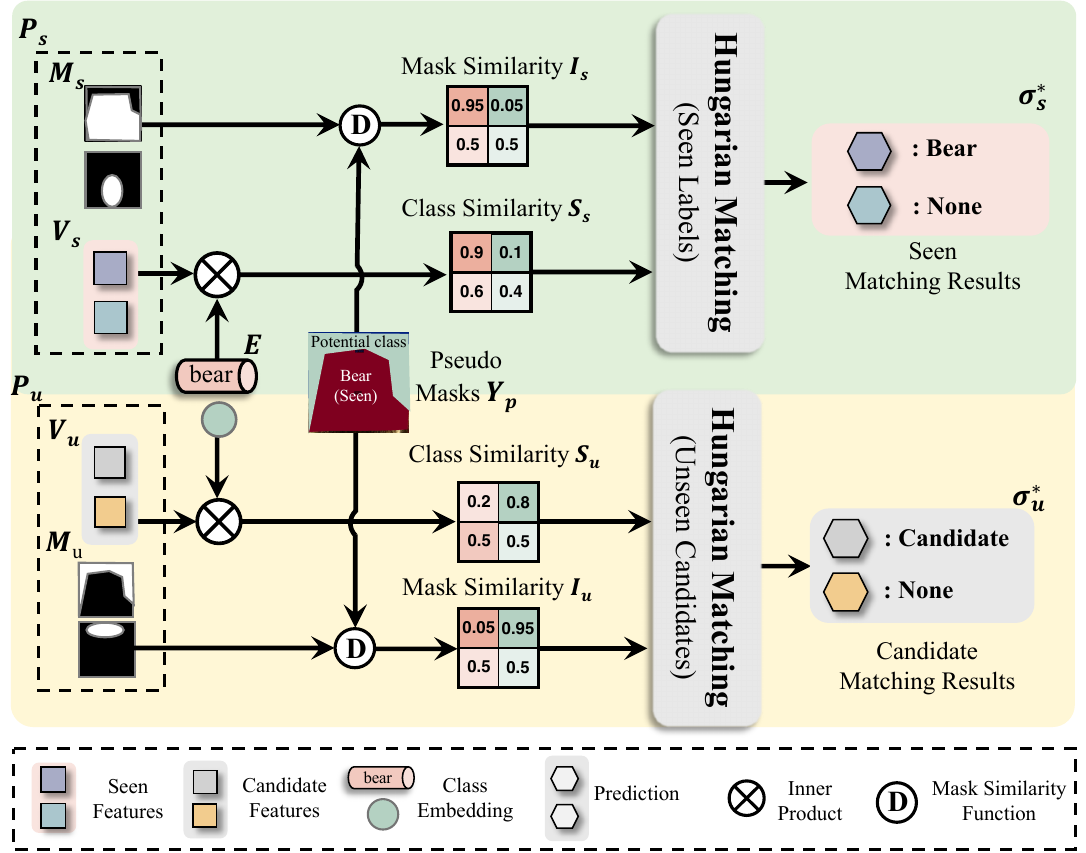}
  \vspace{-0.2in}
  \caption{Overview of SM.}
  \label{fig:SM}
\end{minipage}
\hfill
\begin{minipage}[b]{0.5\linewidth}
  \centering
  \includegraphics[width=\linewidth]{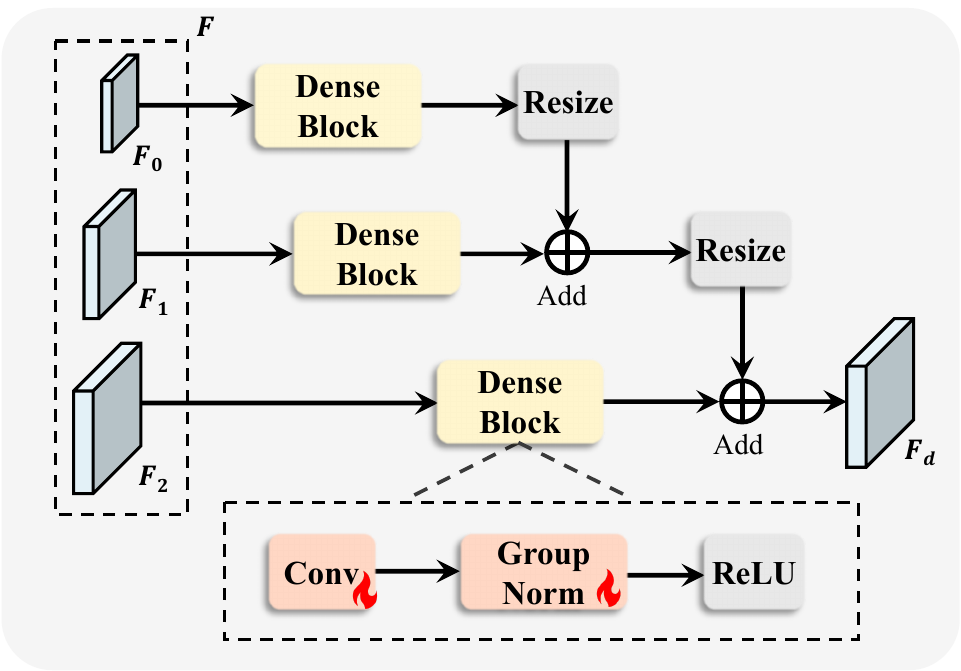}
  \vspace{-0.2in}
  \caption{Overview of MFE.}
  \label{fig:MFE}
\end{minipage}
\vspace{-0.2in}
\end{figure}

To generate the pseudo masks and their corresponding class embeddings for unseen candidates, we use the multi-scale K-means and mask fusion methods from CLIP-ZSS \cite{coach}. Specifically, given an image $\textbf{X}$, we feed it into a frozen CLIP visual encoder to obtain the CLIP dense features $\textbf{O}$. Then, we compute the seed for K-means methods with
\begin{equation}
\label{equa14}
\textbf{G} =\left\{\left.\sum_{u=i}^{i+s-1} \sum_{v=j}^{j+s-1} \frac{\textbf{O}[u, v]}{s^{2}} \right\rvert\, i \in I, j \in J\right\}, 
\end{equation}
where $I = \{0, [s/2], \dots, [H-s]\}$ and $J = \{0, [s/2], \dots, [W-s]\}$ are index sets, $[\cdot]$ denotes rounding, and $s \in S$ is the window size. After we obtain $\textbf{G}$, we merge the masks corresponding to $\textbf{G}$ by the mask fusion algorithm \cite{coach}, which merge the semantically similar masks by the cosine similarity among cluster centroids, to obtain pseudo label $\textbf{Y}_u \in [0,1]^{U \times H \times W}$ for unseen candidates where $U$ indicates the number of unseen candidates. After obtaining $\textbf{Y}_u$, we use $\textbf{Y}_u$ to mask the input images $\textbf{X}$ into $\textbf{X}_m \in \mathbb{R}^{U \times 3 \times H \times W}$. Finally, we feed $\textbf{X}_m$ into the frozen CLIP visual encoder for $\textbf{C}_u \in \mathbb{R}^{U \times C}$ as class embeddings for unseen candidates. Unlike \cite{cluster1} which rely on DINO-ViT \cite{DINO} features, saliency voting, and post-processing (GrabCut), our method directly forms candidate masks without voting or additional refinement. Compared with \cite{zeroguidence}, we do not fix the number of clusters or use image-to-text captioning with CLIP, ensuring no leakage of unseen semantics and reducing computational overhead.



After obtaining $\textbf{Y}_u$ and $\textbf{C}_u$, we apply split matching as illustrated in Fig.~\ref{fig:SM}.
Specifically, we first feed the randomly initialized queries $\textbf{Q}$ into a transformer decoder to obtain the predictions for each query, denoted as $\textbf{P} = \{\textbf{P}_s, \textbf{P}_u\}$.
Here, we denote the predictions for seen and candidate queries as $\textbf{P}_s = \{\textbf{V}_s, \textbf{M}_s\}$ and $\textbf{P}_u = \{\textbf{V}_u, \textbf{M}_u\}$, respectively. Specifically, $\textbf{V}_s \in \mathbb{R}^{K_s \times C}$ and $\textbf{M}_s \in \mathbb{R}^{K_s \times H \times W}$ denote the features after transformer decoder and predicted masks for the $K_s$ seen queries. Similarly, $\textbf{V}_u \in \mathbb{R}^{K_u \times C}$ and $\textbf{M}_u \in \mathbb{R}^{K_u \times H \times W}$ correspond to the predictions for the $K_u$ candidate queries. In both cases, the queries are projected into a $C$-dimensional semantic space. Similar to the conventional hungarian matching, SM also needs to compute the class similarity and mask similarity. 
For class similarity, we concatenate the seen class embeddings $\textbf{A}_s$ (extracted from the CLIP textual encoder) and candidate-class embeddings $\textbf{C}_u$ to form a joint class embedding $\textbf{E} = \mathrm{cat}(\textbf{A}_s, \textbf{C}_u) \in \mathbb{R}^{(N_s + U) \times C}$, where `$\mathrm{cat}$' denotes concatenation. Finally, we separately compute the similarity between the predictions and joint class embeddings for seen and candidate queries, respectively. $\textbf{S}_s = Sigmoid(\textbf{V}_s \cdot \textbf{E}^{\top}), \textbf{S}_u = Sigmoid(\textbf{V}_u \cdot \textbf{E}^{\top})$. Then, for mask similarity, $\textbf{Y}_s$ and $\textbf{Y}_u$ are added to be the total pseudo label $\textbf{Y}_p$ for the mask matching step. The mask similarities are then calculated by comparing these predictions with $\textbf{Y}_p$, $\textbf{I}_s = D(\textbf{M}_s, \textbf{Y}_p), \textbf{I}_u = D(\textbf{M}_u, \textbf{Y}_u)$ where $D$ indicates the function of calculating the similarity between predicted and pseudo masks, \eg, BCE loss. Next, Hungarian matching is then performed independently for seen and candidate queries, ensuring each query group is only matched to its corresponding classes.
\begin{equation}
\sigma_s^* = \arg\min \sum_{i = 0}^{K_s - 1} \mathcal{L}_{match}(\textbf{S}_s^{\sigma(i)}, \textbf{M}_s^{\sigma(i)}, \textbf{E}_i, \textbf{Y}_s^i), \quad
\sigma_u^* = \arg\min \sum_{i = 0}^{K_u - 1} \mathcal{L}_{match}(\textbf{S}_u^{\sigma(i)}, \textbf{M}_u^{\sigma(i)}, \textbf{E}_i, \textbf{Y}_u^i)
\label{eq:SM_joint}
\end{equation}
where $\mathcal{L}_{match}(\textbf{S}^{\sigma}, \textbf{M}^{\sigma}, \textbf{E}, \textbf{Y}) = \mathcal{L}_{cls}(\textbf{S}^{\sigma}, \textbf{E}) + \mathcal{L}_{mask}(\textbf{I}^{\sigma}, \textbf{Y})$, consisting of $\mathcal{L}_{cls}$ and $\mathcal{L}_{mask}$. The classification loss $\mathcal{L}_{cls}$ implemented using focal loss, while the mask loss $\mathcal{L}_{mask}$ is computed using the DICE loss. After computing $\sigma_s^*$ and $\sigma_u^*$, we concatenate them into a unified assignment $\sigma^*$.
$\sigma^*$ is then used to optimize the matching loss $\mathcal{L}^*_{match}$ across all queries,
\begin{equation}
\mathcal{L}^*_{match}(\textbf{S}, \textbf{M}, \textbf{E}, \textbf{Y}) = \mathcal{L}_{cls}(\textbf{S}^{\sigma^*}, \textbf{E}) + \mathcal{L}_{mask}(\textbf{I}^{\sigma^*}, \textbf{Y}_p),
\end{equation}
where $\textbf{S}^{\sigma^*}$ and $\textbf{I}^{\sigma^*}$ are the classification scores and mask similarities after matching with the optimal assignment $\sigma^*$. 
To enhance the discriminability of candidate queries, we introduce a cosine similarity loss: $\mathcal{L}_{cos} = 1 - \mathrm{cos}(\textbf{V}_u', \textbf{C}_u)$, where $\textbf{V}_u'$ is the candidate queries which are assigned with non-igored classes under $\sigma^*_u$, and $\mathrm{cos}(a, b)$ denotes cosine similarity.
The final loss for split matching is $\mathcal{L}_{SM} = \mathcal{L}^*_{match} + \mathcal{L}_{cos}$. Importantly, $\mathcal{L}_{cos}$ is not merely an auxiliary term, but a necessary component of the Split Matching framework, ensuring stable and discriminative alignment of candidate queries in the absence of ground-truth.


\subsection{Multi-scale Feature Enhancement (MFE)} \label{sec:MFE}
Although SM facilitates the adaptation of query-based approaches to zero-shot segmentation, the key and value features in transformer decoders, responsible for associating queries with relevant image regions, remain suboptimal.
Due to the lack of further refinement, the matching performance suffers, ultimately constraining the full potential of the model. To tackle this issue, we propose an Multi-scale Feature Enhancement (MFE) module, as illustrated in Fig. \ref{fig:MFE}, designed to assist in identifying the relevant regions by effectively combining multi-scale features. The MFE leverages hierarchical features extracted by the pixel decoder to provide a comprehensive representation, capturing both fine-grained and global contextual information. The multi-scale outputs of the pixel decoder, $\textbf{F} = \left\{\textbf{F}_0, \textbf{F}_1, \textbf{F}_2\right\}$, represent features at finer resolutions, where $\textbf{F}_0$ has the coarsest resolution and $\textbf{F}_2$ the finest where $\textbf{F}_i \in \mathbb{R}^{C \times (H / r^{2 - i}) \times (W / r^{2 - i})}$, with $r$ denoting the scale factor.

The fusion process begins by refining the lowest-resolution feature map $\textbf{F}_0$ using a dense block with a convolution, group normalization \cite{groupnorm}, and ReLU activation. This block strengthens local spatial cues and prepares $\textbf{F}_0$ for integration with finer-scale features. The refined $\textbf{F}_0$ is resized to match $\textbf{F}_1$, which is similarly refined, and the two are fused via element-wise addition to combine coarse and mid-level semantics. The fused representation is then resized to match $\textbf{F}_2$, the highest-resolution map, which is also refined through its dense block. Finally, the fused $\textbf{F}_0$ and $\textbf{F}_1$ are merged with refined $\textbf{F}_2$ to produce the unified feature map $\textbf{F}_d$, which integrates fine-grained details and contextual cues across scales. We optimize $\textbf{F}_d$ with $\mathcal{L}_{MFE} = \mathcal{L}_{ce}(\textbf{F}_d, \textbf{Y}_p) + \mathcal{L}_{focal}(\textbf{F}_d, \textbf{Y}_p) + \mathcal{L}_{global}(\textbf{F}_d, \textbf{C}_g)$, where $\mathcal{L}_{ce}$, $\mathcal{L}_{focal}$, $\mathcal{L}_{global}$ denote cross entropy, focal loss \cite{focalloss} and global loss in \cite{coach}.





\subsection{Training and Inference} \label{sec:train}
The total loss is $\mathcal{L} = \mathcal{L}_{SM} + \mathcal{L}_{MFE}$.
During inference, we exclude the MFE module and utilize only the backbone and transformer decoder and propose a Random Query (RQ) strategy. Specifically, we feed the trained seen and candidate queries, along with new randomly initialized queries $\mathbf{Q}_r$, into the Transformer decoder. These queries collectively interact with the visual features to generate segmentation masks, where the $\mathbf{Q}_r$ serve to enrich query diversity and improve coverage of unannotated regions.
Unlike the trained queries, $\mathbf{Q}_r$ are not supervised during training and are introduced only at inference time to probe unannotated or ambiguous regions in the image. By increasing the density and diversity of queries in the feature space, $\mathbf{Q}_r$ enhance the model’s ability to explore underrepresented regions that might correspond to unseen or latent objects. Importantly, $\mathbf{Q}_r$ act as complementary probes that are not biased by learned semantic categories, enabling the model to capture alternative activation patterns and recover instances that trained queries might overlook. More detailed inference process and the role of $\mathbf{Q}_r$ are illustrated in the \textbf{\textit{Supplementary Materials}}.



\section{Experiments}
\noindent{\bf Dataset.} We conduct experiments on the widely-used benchmark COCO-Stuff \cite{coco} and PASCAL VOC \cite{voc}, focusing on the task of zero-shot semantic segmentation (ZSS). We adopt the same seen and unseen class splits as in prior works \cite{zegformer,zegclip,maskclip}. Specifically, COCO-Stuff consists of a total of 171 classes with 156 seen and 15 unseen classes according to the standard protocol. The dataset includes 118,287 images for training and 5,000 images for testing. PASCAL VOC contains 10,582 images for training and 1,449 images for validation, including 15 seen and 5 unseen classes.

\begin{table*}[t]

\setlength{\tabcolsep}{23pt}
\resizebox{\linewidth}{!}{
\begin{tabular}{c|ccc|ccc}
\toprule
\multirow{2}{*}{Models} & \multicolumn{3}{c|}{PASCAL VOC}               & \multicolumn{3}{c}{COCO-Stuff}                \\ \cmidrule{2-7} 
                        & \textbf{hIoU} & \textbf{sIoU} & \textbf{uIoU} & \textbf{hIoU} & \textbf{sIoU} & \textbf{uIoU} \\ \midrule
SPNet \cite{spnet}                  & 26.1          & 78.0          & 15.6          & 14.0          & 35.2          & 8.7           \\
ZS3 \cite{zs3}                     & 28.7          & 77.3          & 17.7          & 15.0          & 34.7          & 9.5           \\
CaGNet \cite{cagnet}                 & 39.7          & 78.4          & 26.6          & 18.2          & 33.5          & 12.2          \\
SIGN \cite{sign}                   & 41.7          & 75.4          & 28.9          & 20.9          & 32.3          & 15.5          \\
Joint \cite{joint}                   & 45.9          & 77.7          & 32.5          & -             & -             & -             \\
ZegFormer \cite{zegformer}             & 73.3          & 86.4          & 63.6          & 34.8          & 36.6          & 33.2          \\
Zzseg \cite{simplebaseline}                  & 77.5          & 83.5          & 72.5          & 37.8          & 39.3          & 36.3          \\
DeOP \cite{DeOP}                   & 80.8          & 88.2          & 74.6          & 38.2          & 38.0          & 38.4          \\
ZegCLIP \cite{zegclip}                & 84.3          & {\ul 91.9}    & 77.8          & 40.8          & 40.2          & {\ul 41.4}    \\
OTSeg \cite{otseg}                   & 84.5          & \textbf{92.1} & {\ul 78.1}    & {\ul 41.4}    & {\ul 41.4}    & {\ul 41.4}    \\ \midrule
Ours                    & \textbf{85.3} & 87.7          & \textbf{83.1} & \textbf{42.5} & \textbf{42.6} & \textbf{42.4} \\ \bottomrule
\end{tabular}
}
\caption{Comparison with others. \textbf{Bold} and \underline{underline} indicates the best and the second-best.}
\label{tab:sota}
\vspace{-0.1in}
\end{table*}

\begin{table}[t]
\centering
\begin{minipage}[t]{0.3\linewidth}
\centering
\setlength{\tabcolsep}{5pt}
\resizebox{\linewidth}{!}{
\begin{tabular}{lccc}
\toprule
Method & hIoU & sIoU & uIoU \\ \midrule
Baseline & 24.6 & 31.8 & 20.0 \\
+ SM & 33.3 & 36.4 & 30.8 \\
+ SM + MFE & 36.3 & 36.8 & 35.8 \\
+ SM + MFE + RQ & \textbf{36.6} & \textbf{36.8} & \textbf{36.4} \\
\bottomrule
\end{tabular}
}
\caption{Ablation on the proposed module.}
\label{tab:module_ablation}
\end{minipage}
\hfill
\begin{minipage}[t]{0.34\linewidth}
\centering
\setlength{\tabcolsep}{8pt}
\resizebox{\linewidth}{!}{
\begin{tabular}{lccc}
\toprule
Method & hIoU & sIoU & uIoU \\ \midrule
$\mathbf{F}$ & 33.8 & 35.8 & 32.1 \\
$\mathbf{F}$ + MLP & 30.4 & 36.5  & 26.1 \\
$\mathbf{C}_u$ & \textbf{36.6} & \textbf{36.8} & \textbf{36.4} \\
\bottomrule
\end{tabular}
}
\caption{Ablation on the candidate class embedding.}
\label{tab:class_embedding}
\end{minipage}
\hfill
\begin{minipage}[t]{0.34\linewidth}
\centering
\setlength{\tabcolsep}{5pt}
\resizebox{\linewidth}{!}{
\begin{tabular}{lccc}
\toprule
Structure of MFE & hIoU & sIoU & uIoU \\ \midrule
No Norm & 35.8 & 36.2 & 35.3 \\
BN &  36.0 & 36.5 & 35.6 \\
GN & \textbf{36.6} & \textbf{36.8} & \textbf{36.4} \\
\bottomrule
\end{tabular}
}
\caption{Ablation on the structure of MFE.}
\label{tab:norm_ablation}
\end{minipage}
\vspace{-0.2in}
\end{table}

\noindent{\bf Implementation Details.} The CLIP model applied in our method is based on the ViT-B/16 model, and the channel of the output text features is 512. All the experiments are conducted on 8 V100 GPUs, and the batch size is set to 16 for both datasets. The iterations are set to 20K and 80K for PASCAL VOC and COCO-Stuff. $\mathcal{L}_{cls}$ in $\mathcal{L}_{match}$ is focal loss \cite{focalloss} and the $\mathcal{L}_{mask}$ is a combination of IoU loss and DICE loss \cite{mask2former} in $\mathcal{L}_{match}$. We choose Mask2Former \cite{mask2former} with ResNet101 as the backbone with all other hyperparameters unchanged. 50 unseen and 50 random queries are added during inference. We apply the harmonic mean IoU (hIoU) following previous works \cite{zegclip} where $ \small{hIoU} = \frac{2 \cdot sIoU \cdot uIoU}{sIoU + uIoU}$ as the metric
where $sIoU$ and $uIoU$ indicate the mIoU (mean intersection over union) of the seen classes and unseen classes, respectively. More details are in the \textbf{\textit{Supplementary Materials}}.

\subsection{Comparison with State-of-the-art Methods}
Table \ref{tab:sota} shows that our method achieves state-of-the-art performance, outperforming existing methods in terms of overall hIoU and uIoU. Specifically, we obtain the highest uIoU on PASCAL VOC (83.1\%), which is a significant margin over ZegCLIP (77.8\%) and OT-Seg (78.1\%), demonstrating stronger generalization to unseen classes. Importantly, our method also achieves the best hIoU on both datasets, indicating more balanced segmentation. Although our sIoU on VOC is slightly lower than transformer-based counterparts, this reflects our model’s ability to mitigate seen-class bias.



\begin{figure*}[t]
\centering
\begin{minipage}[t]{0.48\linewidth}
\centering
\includegraphics[width=\linewidth]{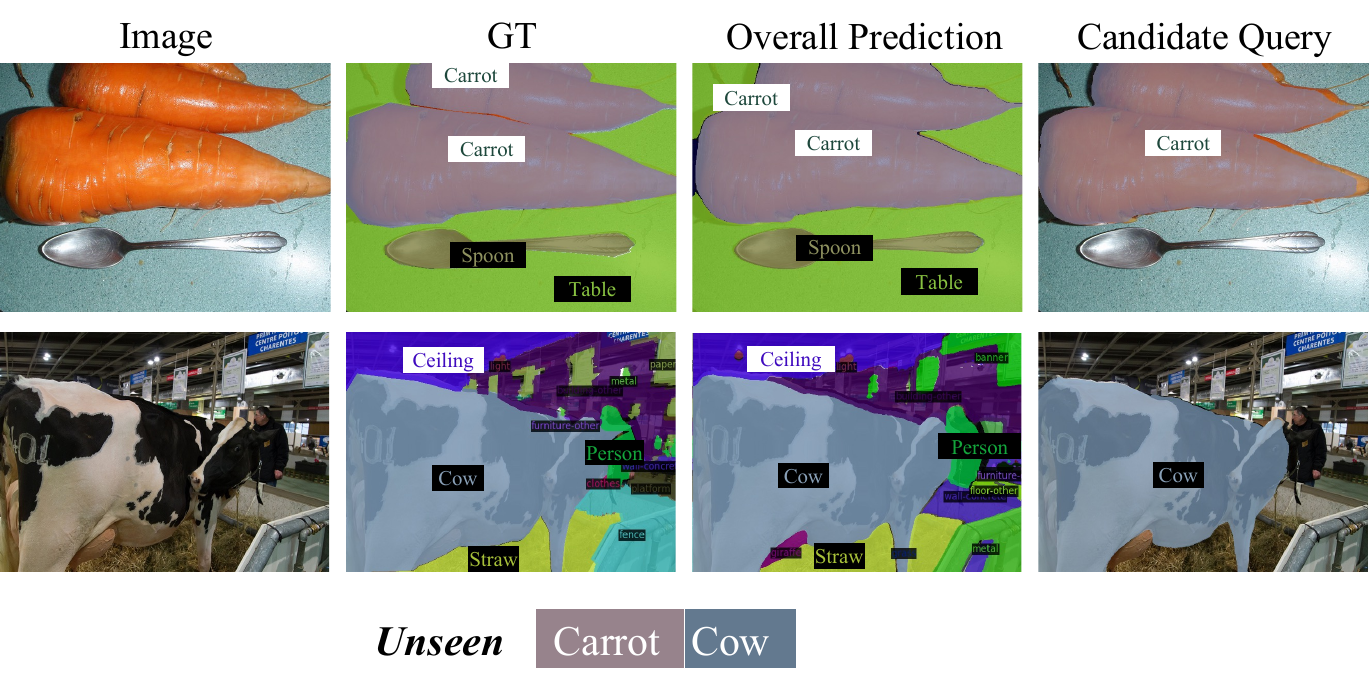}
\vspace{-0.3in}
\caption{Candidate query predictions visualization, with each row displaying images, GT, overall and candidate query predictions.}
\label{fig:unseen query}
\end{minipage}
\hfill
\begin{minipage}[t]{0.48\linewidth}
\centering
\includegraphics[width=\linewidth]{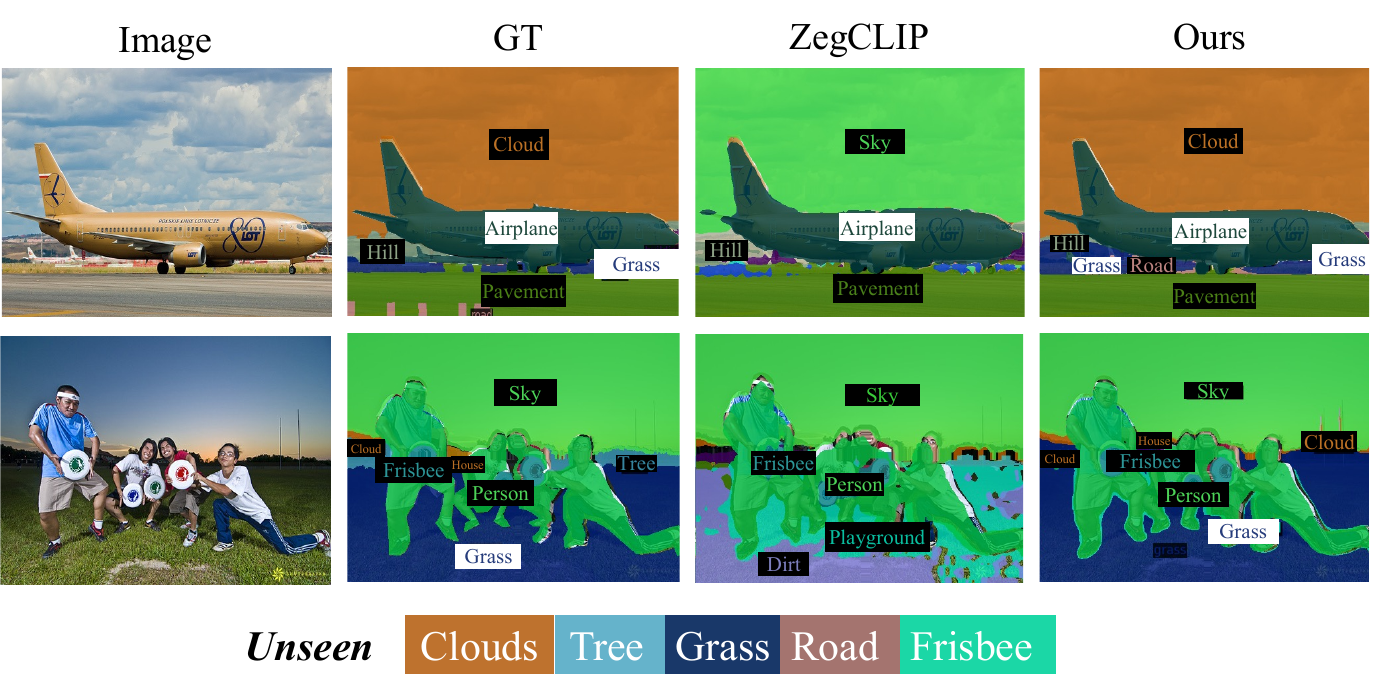}
\vspace{-0.3in}
\caption{Visualization on predictions where each line shows the image, ground truth, ZegCLIP's prediction, and ours.}
\label{fig:vis}
\end{minipage}
\vspace{-0.2in}
\end{figure*}

\subsection{Ablation Studies} \label{sec:ablation}

To evaluate the effectiveness of our method, we conduct ablation studies on COCO-Stuff for 40K iterations using ResNet-50 as the backbone, with all hyperparameters unchanged.

\noindent \textbf{Ablations on Proposed Modules.}  
Table \ref{tab:module_ablation} summarizes the contributions of each proposed module. The baseline achieves suboptimal performance, with lower unseen IoU (uIoU) indicating limited generalization. Adding the Split Matching (SM) module significantly improves the model’s ability to capture unseen classes, as reflected in higher uIoU. The Multi-scale Feature Enhancement (MFE) further boosts the model's performance by enhancing the interaction between queries and features. Finally, the inclusion of random queries leads to the best results across all metrics, demonstrating our contribution.

\noindent \textbf{Ablations on Unseen Class Embedding.}  
Table \ref{tab:class_embedding} presents an ablation study on the design of unseen class embeddings $\mathbf{C}_u$. Using raw dense features $\mathbf{F}$ from the backbone provides a reasonable baseline (uIoU: 32.1), while adding an MLP projection degrades performance, likely due to semantic distortion. In contrast, using CLS tokens from pseudo-masked regions yields the best results (uIoU: 36.4), suggesting that region-level CLS tokens better preserve semantic alignment. This also enables direct concatenation with seen-class text embeddings, maintaining cross-modal consistency without modality mismatch.

\noindent \textbf{Ablations on structure of MFE.} Table \ref{tab:norm_ablation} presents the ablation study on the structure of the MFE module by comparing different normalization strategies: No Normalization (No Norm), Batch Normalization (BN), and Group Normalization (GN).
Among the three, GN achieves the best performance across all metrics, yielding the highest hIoU, sIoU, and uIoU.

\subsection{Qualitative Analysis}

\begin{figure}[t]
\centering
\begin{minipage}[b]{0.48\linewidth}
    \centering
    \includegraphics[width=0.85\linewidth]{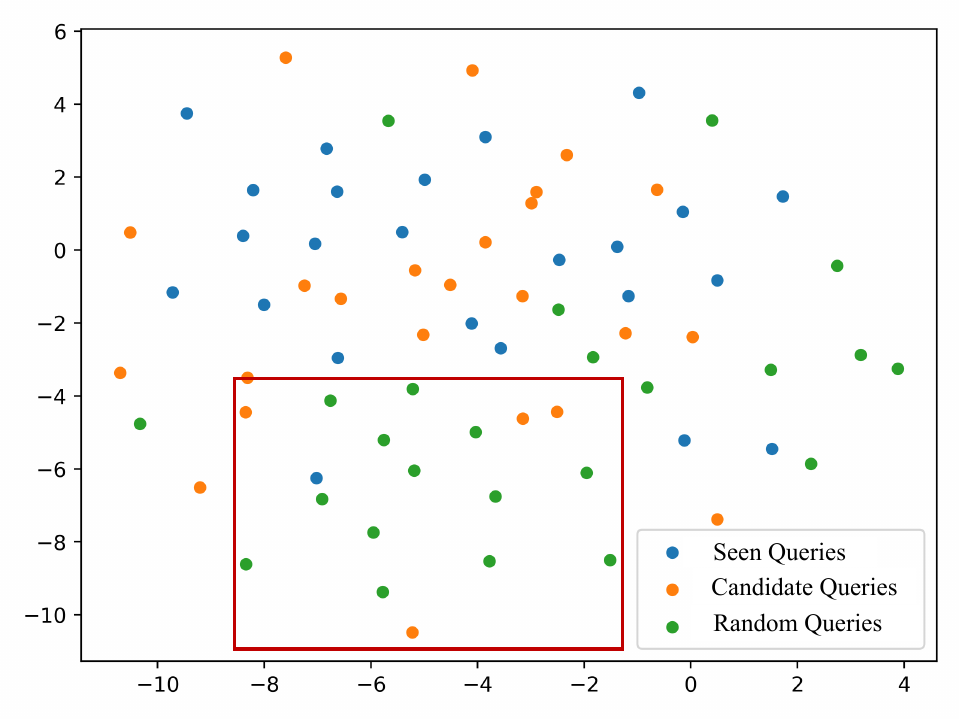}
    \vspace{-0.1in}
    \caption{T-SNE visualization among all three types of queries.}
    \label{fig:query_dis}
\end{minipage}
\hfill
\begin{minipage}[b]{0.48\linewidth}
    \centering
    \includegraphics[width=0.85\linewidth]{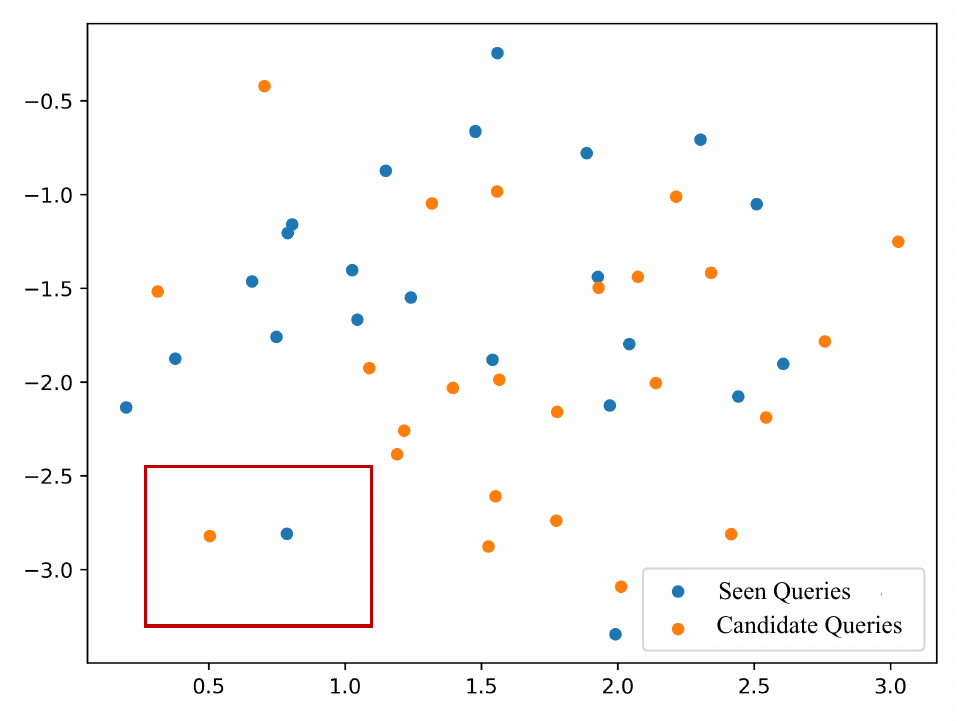}
    \vspace{-0.1in}
    \caption{T-SNE visualization of seen and candidate queries.}
    \label{fig:query_dis_su}
\end{minipage}
\label{fig:tsne}
\vspace{-0.2in}
\end{figure}

\noindent \textbf{Visualization of candidate queries.} Fig.~\ref{fig:unseen query} visualizes the predictions of candidate queries. Notably, these queries successfully activate on previously unannotated regions, enabling the model to localize unseen classes such as \textit{carrot} and \textit{cow}. This demonstrates that candidate queries can effectively discover latent classes and assign semantically correct class labels, even without explicit supervision. \textbf{Prediction Visualzation.}  Each row in Fig. \ref{fig:vis} shows the input image, ground truth, ZegCLIP’s prediction, and ours. Our method successfully segments unseen classes such as “clouds”, “bushes”, and “playingfield”, which are missed or mislabeled by ZegCLIP. Notably, in both examples, the unseen class “clouds” is correctly identified by our model, demonstrating better generalization to unseen concepts. \textbf{Visualization of query distribution.} Fig. \ref{fig:query_dis} shows seen (blue), candidate (orange), and random (green) queries in the embedding space. Random queries occupy a concentrated region (red box) that complements areas underrepresented by seen and candidate queries, improving coverage of unseen classes. Fig. \ref{fig:query_dis_su} appears sparser in the lower-left region, explaining the performance gain from random queries by increasing feature density.

\section{Conclusion}
In this paper, we propose \textbf{Split Matching} (SM), a novel decoupled assignment strategy tailored for query-based models in ZSS. By separating queries into seen and candidate groups and optimizing them with respect to annotated and unannotated regions, SM effectively mitigates the seen-class bias caused by incomplete supervision. To further facilitate the discovery of unseen classes, we leverage CLIP-derived pseudo masks and region-level embeddings, and introduce a \textbf{Multi-scale Feature Enhancement} (MFE) module to refine spatial representations. Additionally, we incorporate a \textbf{Random Query} (RQ) strategy during inference to improve query diversity and coverage of unannotated regions. Extensive experiments on standard ZSS benchmarks demonstrate that our approach achieves state-of-the-art results.

\section*{Acknowledgment}
Support for this work was given by the Toyota Motor Corporation (TMC) and JSPS KAKENHI Grant Number 23K28164 and JST CREST Grant Number JPMJCR22D1. However, note that this paper solely reflects the opinions and conclusions of its authors and not TMC or any other Toyota entity. Computations are done on the supercomputer “Flow” at the Information Technology Center, Nagoya University.

\bibliography{egbib}
\end{document}